\documentclass{article}

\usepackage[final]{nips_2017}

\usepackage[utf8]{inputenc} 
\usepackage[T1]{fontenc}    
\usepackage{hyperref}       
\usepackage{url}            
\usepackage{booktabs}       
\usepackage{amsfonts}       
\usepackage{nicefrac}       
\usepackage{microtype}      
\usepackage{wrapfig}

\usepackage{amsmath,amssymb}
\usepackage{subfigure}
\usepackage{graphicx}
\usepackage{multirow}
\usepackage{mathrsfs}
\usepackage{algorithm}
\usepackage[noend]{algpseudocode}
\makeatletter
\def\BState{\State\hskip-\ALG@thistlm}
\makeatother

\DeclareMathOperator*{\argmax}{arg\,max} 

\usepackage{listings}
\usepackage{caption}

\title{Personalizing a Dialogue System with Transfer Reinforcement Learning}

\author{
Kaixiang Mo$^\dag$, Yu Zhang$^\dag$, Shuangyin Li$^\dag$, Jiajun Li$^\ddag$, Qiang Yang$^\dag$  \\
The Hong Kong University of Science and Technology, Hong Kong, China \\
\texttt{$^\dag$\{kxmo, zhangyu, shuangyinli, qyang\}@cse.ust.hk $^\ddag$\{jiajun.li\}@alumni.ust.hk}\\
}

\begin{document}

\maketitle

\begin{abstract}
It is difficult to train a personalized task-oriented dialogue system because the data collected from each individual is often insufficient.
Personalized dialogue systems trained on a small dataset is likely to overfit and make it difficult to adapt to different user needs.
One way to solve this problem is to consider a collection of multiple users as a source domain and an individual user as a target domain, and to perform transfer learning from the source to the target domain.
By following this idea, we propose a PErsonalized Task-oriented diALogue (PETAL) system, a transfer learning framework based on POMDP to construct a personalized dialogue system. The PETAL system first learns common dialogue knowledge from the source domain and then adapts this knowledge to the target domain. The proposed PETAL system can avoid the negative transfer problem by considering differences between source and target users in a personalized Q-function. 
Experimental results on a real-world coffee-shopping data and simulation data show that the proposed PETAL system can learn different optimal policies for different users, and thus effectively improve the dialogue quality under the personalized setting.
\end{abstract}

\section{Introduction}
Dialogue systems can be classified into two classes: open domain dialogue systems~\cite{ritter2011data,galley2015deltableu,serban2015hierarchical,li2016deep,mou2016sequence} and task-oriented dialogue systems~\cite{levin1997learning,young2013pomdp,wen2015semantically,wen2016network,williams2016end}.
Open domain dialogue systems do not limit the dialogue topic to a specific domain, and typically do not have a clear dialogue goal. Task-oriented dialogue systems aim to solve a specific task via dialogues. In this paper we focus on the dialogue systems which aim to assist users to finish a task such as ordering a cup of coffee.
Personalized task-oriented dialogue systems aim to help a user complete a dialogue task better and faster than non-personalized dialogue systems. Personalized dialogue systems can learn about the preferences and habits of a user during interactions with the user, and then utilize these personalized information to speed up the conversation process.
Personalized dialogue systems could be categorized into rule-based dialogue systems~\cite{thompson2004personalized,kim2014acquisition,bang2015example} and learning-based dialogue systems~\cite{casanueva2015knowledge,genevay2016transfer}.
In rule-based personalized dialogue systems, the dialogue state, system speech act and user speech act are predefined by developers, hence it is difficult for us to use this system when the dialogue state and the speech act are hard to define manually. Learning-based personalized dialogue systems could learn states and actions from training data without requiring explicit rules designed by developers. 

However,
it is difficult to train a personalized task-oriented dialogue system because the data collected from each individual is often insufficient. A personalized dialogue system trained on a small dataset is likely to fail on unseen but common dialogue cases due to over-fitting.
One solution is to consider a collection of multiple users as a source domain and an individual user as a target domain, and transfer common dialogue knowledges to the target domain.
When transferring dialogue knowledge, the challenge lies in the difference between the source and target domains.
Some works~\cite{casanueva2015knowledge,genevay2016transfer} have been proposed to transfer dialogue knowledge among similar users, but they did not model the difference between different users, which might harm the performance in the target domain.

\begin{wrapfigure}{r}{0.45\textwidth}
\vskip -0.1in
\begin{minipage}{0.45\textwidth}
\begin{figure}[H]
\centering
\includegraphics[width=\textwidth]{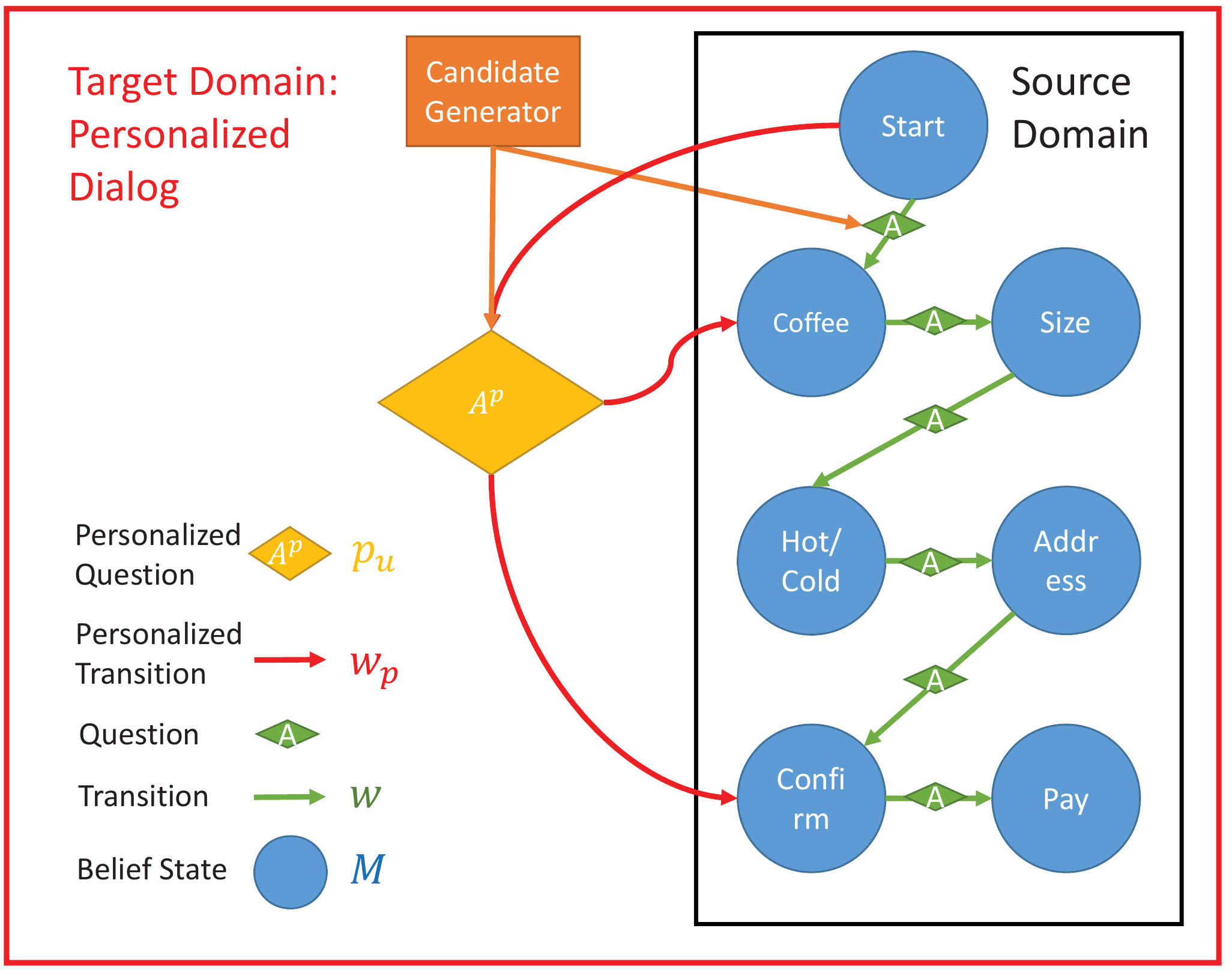}
\caption{The flowchart of the proposed PETAL system on the coffee-ordering task.} \label{fig:Flowchart}
\end{figure}
\end{minipage}
\vskip -0.1in
\end{wrapfigure}

In this paper, we propose a PErsonalized Task-oriented diALogue (PETAL) system, which is a transfer learning framework based on the POMDP for learning a personalized dialogue system. The PETAL system first learns common dialogue knowledge from the source domain and then adapts this knowledge to the target user.
To achieve this goal, the PETAL system models personalized policies with a personalized Q-function defined as the expected cumulative general reward plus expected cumulative personal reward. The personalized Q-function can model differences between the source and target users, and thus can avoid the negative transfer problem brought by differences between source and target users. The flowchart of the PETAL system on the coffee-ordering task is shown in Figure~\ref{fig:Flowchart}.
Experimental results on a real-world coffee-shopping dataset and simulation data show that the proposed PETAL system can choose different optimal actions for different users, and thus effectively improves the dialogue quality under the personalized setting.

Our contributions are three-fold: Firstly, we tackle the problem of learning common dialogue knowledge from the source domain and adapting to the target user in a personalized dialogue system. In multi-turn dialogue systems, learning optimal responses in different situations is a non-trivial problem. One naïve policy is to always choose previous seen sentences, but it is not necessarily optimal. For example in the online coffee ordering task, such naïve policy could incur many logical mistakes such as asking repeated questions and confirming the order before the user finishes ordering.
Secondly, we propose a transfer learning framework on the POMDP to model the preferences of different users.
Unlike existing methods, the proposed PETAL system does not require a manually-defined ground truth state space and it can model the personalized future expected reward. Finally, we demonstrate the effectiveness of the PETAL system on a real-world dialogue dataset as well as simulation data.
\section{Related Works}
\label{RelatedWork}
Personalized dialogue systems could be categorized into rule-based dialogue systems and learning-based dialogue systems.
For rule-based personalized dialogue systems, Thompson et al.~\cite{thompson2004personalized} propose an interactive system where users can choose a place via an interactive conversational process and the system could learn user preference to improve future conversations. Personalization frameworks proposed in ~\cite{kim2014acquisition,bang2015example} extract and utilize user-related facts (triples), and then generate responses by applying predefined templates to these facts.
Different from rule-based personalized dialogue systems, learning-based personalized dialogue systems can learn states and actions from training data without requiring explicit rules.
Casanueva et al.~\cite{casanueva2015knowledge} propose to initialize personalized dialogue systems for a target speaker with data from similar speakers in the source domain to improve the performance for the target speaker. This work requires a predefined user similarity metric to select similar source users, and when the selected similar users are different from the target user, the performance for the target user will degrade.
Genevay and Laroche~\cite{genevay2016transfer} propose to select and transfer an optimized policy from source users to a target user by using a multi-armed stochastic bandit algorithm which does not require a predefined user similarity measure.
However, this method has a high complexity since for each target user, it requires $n^2$ bandit selection operations where $n$ is the number of source users.
Moreover, similar to \cite{casanueva2015knowledge}, the differences between selected source users and the target user will deteriorate the performance. 
Different from these works, the proposed method does not assume the predefined dialogue states and system speech acts required by the rule-based systems, and it explicitly models the differences between users.

Transfer learning~\cite{taylor2009transfer,pan2010survey,tan2014multi,tan2015transitive,wei2016transfer} has been applied to other tasks in dialogue systems. Gasic et al.~\cite{gavsic2013pomdp} uses transfer learning to extend a dialogue system to include a previously unseen concept. Gasic et al.~\cite{gavsic2014incremental} propose an incremental scheme to adapt an existing dialogue management system to an extended domain. These two works transfer parameters in the policy of the source domain as a prior to the target domain. However, these two models do not deal with multiple source domains and they do not have explicit personalized mechanisms for different users. As a consequence, negative transfer might occur when the differences between users are large. In contrast, the proposed method has an explicit personalization mechanism and can alleviate negative transfer.

In argumentation agents, there are some works \cite{hiraoka2014reinforcement,rosenfeld2016strategical,rosenfeld2016providing} which study personalized dialogue system. However, these works, which aim to influence users' goal, have different motivations from ours and their formulations are totally different from ours.


\section{PETAL: A Framework for Personalized Dialogue Management}

In this section, we introduce the proposed PETAL system. Here we use PETAL to denote both the proposed framework and the proposed algorithm.

\subsection{Problem Setting}
Matrices are denoted in bold capital case, row vectors are in bold lower case and scalars are in lower case. The text in the dialogues, denoted in curlicue, is represented by the the bag-of-words assumption. Each of the bag-of-words representations is a vector in which each entry has a binary value.

Since the current state of the dialogue is not observable and the ground truth dialogue states are assumed to be unknown, we formulate the dialogue as a POMDP, which is defined as 7-tuple $\{S, A, O, P, R, Z, \gamma \}$, where $S$ denotes the hidden unobservable states, $A$ denotes the replies of the agent, $O$ denotes users' utterances, $P$ is the state transition probability function, $R$ is the reward function, $Z$ is the observation function, and $\gamma \in [0,1]$ is the discounted factor. In the $i$-th turn of a dialogue with a user $u$, $S_i^u$ is the hidden conversation state, $\mathcal{O}_i^u$ is the user utterance, $\mathcal{A}_i^u$ is the reply of the agent, and $r_i^u$ is the reward. In the $i$-th turn, we only observe $\mathcal{O}_i^u$, $\mathcal{A}_i^u$ and $r_i^u$. We define $\mathbf{b}_i^u$ as the belief state vector, which represents the probability distribution of unobserved $S_i^u$.
Unlike previous work, we do not assume that the underlying ground truth state space $S$ is provided. Instead we propose to learn a function to map the dialogue history $\mathcal{H}_i^{u}=\{ \{\mathcal{O}_k^{u},\mathcal{A}_k^{u}\}_{k=0}^{i-1}, \mathcal{O}_i^{u} \}$ to a compact belief state vector $\mathbf{b}_i^u$.


\noindent The inputs for this problem include
\begin{enumerate}
\item Abundant dialogue data $ \{\{\mathcal{O}_i^{u_{s}}, \mathcal{A}_i^{u_{s}}\}_{i=0}^T\}$ of source customers $u_{s}$.
\item A few dialogue data $ \{\{\mathcal{O}_i^{u_{t}}, \mathcal{A}_i^{u_{t}}\}_{i=0}^T\}$ of the target customer $u_{t}$.
\end{enumerate}
\noindent The expected output is
\begin{enumerate}
\item A policy $\pi_{u_{t}}$ for target user.
\end{enumerate}

\subsection{The Framework}

In order to solve the problem, we aim to find a policy $\pi_{u_{t}}$ for the target user, which could choose an appropriate action $\mathcal{A}_i^{u_t}$ at the $i$-th turn based on current dialogue history $\mathcal{H}_i^{u_t}$, to maximize the cumulative reward defined as
$ \pi_{u_{t}} = \argmax_\pi \mathbb{E} \left[ \sum_{k=0}^{\infty} \gamma^k r_{t+k+1}^{u_t} \right]$.

To model belief states, we introduce a state projection matrix $\mathbf{M}$ to map dialogue history $\mathcal{H}_i^u$ to belief state $\mathbf{b}_i^u$, i.e., $ {\mathbf{b}}_i^u = f(\mathcal{H}_i^u;{\mathbf{M}})$.

The Q-function is defined as the expected cumulative reward according to policy $\pi_u$ by starting from belief state $\mathbf{b}_i^u$ and taking action $\mathcal{A}_i^u$ as
\begin{align*}
Q^{\pi_u}(\mathcal{H}_i^u, \mathcal{A}_i^u) = E_{\pi} \left[\sum_{k=0}^{\infty} \gamma^k r_{t+k+1}^u |\mathcal{H}_i^u, \mathcal{A}_i^u\right].
\end{align*}
We choose value-based approaches because there is usually a small number of training data in the target domain, while policy-based approaches, which generate responses word by word, require a lot of training data.

In order to build a personalized dialogue system for the target user, we need to learn a personalized Q-function $Q^{\pi_{u_{t}}}$ for this user. However, since the training data $\{\{\mathcal{O}_i^{u_{t}}, \mathcal{A}_i^{u_{t}}\}_i^T\}$ for the target user $u_{t}$ is very limited, we can hardly estimate the personalized Q-function $Q^{\pi_{u_{t}}}$. In order to learn an accurate $Q^{\pi_{u_{t}}}$, we can transfer common dialogue knowledge from the source domain, which has a lot of data from many other users $\{\{\mathcal{O}_{i}^{u_{s}}, \mathcal{A}_i^{u_{s}}\}_{i=0}^T\}$. However, different users may have different preferences, hence directly using the data from source users would bring negative effects. We propose to model the personalized Q-function as a general Q-function $Q_g$ plus a personal one $Q_p$:
\begin{align*}
Q^{{\pi_{u}}}(\mathcal{H}_i^u, \mathcal{A}_i^u) =& Q_g(\mathcal{H}_i^u, \mathcal{A}_i^u ; \mathbf{w}) + Q_p(\mathcal{H}_i^u, \mathcal{A}_i^u ; \mathbf{p}_u, w_p )\\
\approx&\mathbb{E}_{\pi_u}\left [ \sum_{k=0}^{\infty} \gamma^k r^{u,g}_{t+k+1} |\mathcal{H}_i^u, \mathcal{A}_i^u \right] + \mathbb{E}_{\pi_u} \left[\sum_{k=0}^{\infty} \gamma^k r^{u,p}_{t+k+1} |\mathcal{H}_i^u, \mathcal{A}_i^u\right],
\end{align*}
where $r^{u,g}_{t}$ and $r^{u,p}_{t}$ denotes the general and personal rewards for user $u$ at time $t$ respectively, the general Q-function $Q_g(\mathcal{H}_i^u, \mathcal{A}_i^u ; \mathbf{w})$ captures the expected reward related to the general dialogue policy for all users, $\mathbf{w}$ is the set of parameters for the general Q-function and contains a large amount of parameters such that it requires a lot of training data, and the personal Q-function $Q_p(\mathcal{H}_i^u, \mathcal{A}_i^u ; \mathbf{p}_u, w_p )$ captures the expected reward related to the preference of each user.

The proposed framework is based on transfer learning. $\mathbf{M}$, $\mathbf{w}$ and $w_p$ are shared across different users, which could be trained on source domains and then transferred to the target domain.
These parameters contain the common dialogue knowledge, which is independent of users' preferences. Moreover, $\mathbf{p}_u$, which is user-specific, capture the preferences of different users.

\subsection{Parametric Forms for Personalized Q-function}

In this section, we introduce parametric forms for $f(\mathcal{H}_i^u ; {\mathbf{M}})$, $Q_g(\mathcal{H}_i^u, \mathcal{A}_i^u ; \mathbf{w})$ and $Q_p(\mathcal{H}_i^u, \mathcal{A}_i^u ; \mathbf{p}_u, w_p )$ in the personalized Q-function.

Dialogue states are defined as follows. All utterances and replies will be projected into state vectors with a state projection matrix $\mathbf{M}$, where $\mathbf{M}$ is initialized with the word2vec and will be updated in the learning process. ${\mathbf{b}}_i^u = f(\mathcal{H}_i^u ; \mathbf{M})$ maps the dialogue history, $\mathcal{H}_i^u=\{\{\mathcal{O}_{k}^u, \mathcal{A}_{k}^u\}_{k=0}^{i-1}, \mathcal{O}_{i}^u\}$, to a belief state vector.
The belief state vector $\mathbf{b}_i^u$ is defined as
${\mathbf{b}}_i^u = \left[{\mathbf{o}}^{h,u}_{i-1}, {\mathbf{o}}_{i}^u, {\mathbf{a}}^{h,u}_{i-2}, {\mathbf{a}}_{i-1}^u\right]$,
where $\xi=0.8$ is the memory factor to discount historical state vectors at each time step, $\mathbf{o}^{h,u}_i = \sum_{k=0}^i \xi^{i-k} \mathbf{o}_k^u$,
$\mathbf{o}_i^u = \mathcal{O}_{i}^u \mathbf{M}$, $ \mathbf{a}^{h,u}_i = \sum_{k=0}^i \xi^{i-k} \mathbf{a}_k^u$, and $\mathbf{a}_{i-1}^u = \mathcal{A}_{i-1}^u \mathbf{M}$. Based on these definitions, we can see that $\mathbf{o}^{h,u}_i$ represents all previous user utterances, $\mathbf{o}_i^u$ represents the current user utterance, $\mathbf{a}^{h,u}_i$ represents all previous agent replies, and $\mathbf{a}_{i-1}^u$ represents the last agent reply.

In order to model the correlations between entries in $\mathbf{a}_i^u$ and $\mathbf{b}_i^u$, the general Q-function $Q_g(\mathcal{H}_i^u, \mathcal{A}_i^u ; \mathbf{w})$ is defined as
\begin{equation*}
Q_g(\mathcal{H}_i^u, \mathcal{A}_i^u ; \mathbf{w})=\mathbf{a}_i^u\mathbf{W}({\mathbf{b}_i^u})^T,
\end{equation*}
where superscript $^T$ denotes the transpose of a vector or matrix, $\mathbf{W}\in\mathbb{R}^{d\times 4d}$ is a parameter matrix to be learned. Based on the properties of the Kronecker product and operator $\mathrm{vec}(\cdot)$ which transforms a matrix to a vector in a columnwise manner, we can rewrite $Q_g(\mathcal{H}_i^u, \mathcal{A}_i^u ; \mathbf{w})$ as a linear function on $\mathbf{w}=\mathrm{vec}(\mathbf{W})^T\in\mathbb{R}^{4d^2}$: $Q_g(\mathcal{H}_i^u, \mathcal{A}_i^u ; \mathbf{w}) = ({\mathbf{b}}_i^u \otimes {\mathbf{a}}_i^u) {\mathbf{w}}^T$, where $\mathbf{b}_i^u \otimes \mathbf{a}_i^u$ is the Kronecker product of $\mathbf{b}_i^u$ and $\mathbf{a}_i^u$.
In multi-round dialogue systems, there should be different optimal actions in different belief states. The rationale to use the Kronecker product is that the general Q-function should depend on the combination of belief state $\mathbf{b}_i^u$ and action $\mathbf{a}_i^u$, but not independently on $\mathbf{b}_i^u$ and $\mathbf{a}_i^u$.


The personal Q-function learns personalized preference for each user to avoid the negative effect brought by transferring biased dialogue knowledge across users with different preferences. We denote by $C_j$ the set of all possible choices in the $j$-th choice set we want to collect and by $\{c^u_{ij}\}_{j=1}^m$ the choices proposed in the $i$-th agent response $\mathcal{A}_i^u$, where $m$ is the total number of order choices, hence $c^u_{ij}$ is an exact choice in $C_j$. For example, in the coffee-ordering task, $C_1=\{$Latte, Cappuccino,$\ldots\}$ could be the type of coffees and $c^u_{i1}$ could be any coffee in $C_1$. From the user side, $c^u_{ij}$ is just the choice of user $u$ for the $j$-th choice set in the $i$-th dialogue turn. For example, $c^u_{i1}$ could be ``latte'' and $c^u_{i2}$ could be ``iced''.
Based on an assumption that different choice sets are independent of each other, for the $j$-th choice set, the probability of a user $u$ to choose $c^u_{ij}$ follows a categorical distribution $\mathcal{C}(c^u_{ij};\mathbf{p}^u_{j})=p^u_{j,c^u_{ij}}$ where $|C_j|$ denotes the cardinality of a set,  $\mathbf{p}^u_{j} \in \mathbb{R}^{|C_j|}$, and $p^u_{j,k}$ denotes the $k$-th entry in $\mathbf{p}^u_{j}$. Hence the personal Q-function for user $u$ is formulated as
\begin{equation*}
Q_p(\mathcal{H}_i^u, \mathcal{A}_i^u ; \mathbf{p}_u, w_p ) = w_p \sum_{j=1}^m \mathcal{C}(c^u_{ij} ; {\mathbf{p}}_{uj}) \delta( C_j ,\mathcal{H}_i^u),
\end{equation*}
where the personal preference $\mathbf{p}_{u}=\{\mathbf{p}_{uj}\}_{j=1}^m$ for user $u$ is learned from training process, $\delta(C_j, \mathcal{H}_i^u)$ equals 1 if the user has not yet made a choice about $C_j$ in the dialogue history $\mathcal{H}_i^u$ and 0 otherwise.
$\delta(C_j, \mathcal{H}_i^u)$ implies whether the system will receive a personal reward in the rest of the dialogue, as the Q-function models the cumulative future reward.
Here $w_p$ controls the importance of the personalized reward and it is learned from data. When $w_p$ is close to zero, the Q-function will depend on the general dialogue policy. Note that $\sum_{j=1}^m \mathcal{C}( c^u_{ij} | {\mathbf{p}}_{uj}) \delta (C_j, \mathcal{H}_i^u)$ is 0 if we know nothing about the user, or $\mathcal{A}_i^u$ does not show any personal preference of user $u$. Because the vocabulary of choices is much smaller than the whole vocabulary, we can estimate the personal preference parameters $\mathbf{p}_u$ with a few dialogue data $\{\{\mathcal{O}_i^{u_{t}}, \mathcal{A}_i^{u_{t}}\}_i^T\}$ from the target user. 

By combining the general and personal Q-functions, the personalized Q-function can finally be defined as
\begin{equation*}
Q^{{\pi_{u}}}(\mathcal{H}_i^u, \mathcal{A}_i^u) = ({\mathbf{b}}_i^u \otimes {\mathbf{a}}_i^u) {\mathbf{w}}^T + w_p \sum_{j=1}^m \mathcal{C}(c^u_{ij} | {\mathbf{p}}_{uj}) \delta( C_j, \mathcal{H}_i^u).
\end{equation*}
Here $\mathbf{M}, \mathbf{w}, w_p$ are shared across different users, which could be trained on the source domains and then transferred to the target domain.

\subsection{Reward}
\label{sec:reward}

The total reward is the sum of general reward and personal reward, which can be defined as follows:
\begin{enumerate}
\item A personal reward $r^{u,p}$ of $0.3$ will be received when the user confirms the suggestion of the agent, and a negative reward of $-0.2$ will be received if the user declines the suggestion by the agent. This is related to the personal information of the user. For example, the user could confirm the address suggested by the agent.
\item A general reward $r^{u,g}$ of $0.1$ will be received when the user provides the information about each $c_j$.
\item A general reward $r^{u,g}$ of $1.0$ will be received when the user proceeds with payment.
\item A general reward $r^{u,g}$ of $-0.05$ will be received by the agent for each dialogue turn to encourage shorter dialogue, $-0.2$ will be received by the agent if it is generates non-logical responses such as asking repeated questions.
\end{enumerate}
Note that the personal reward could not be distinguished from the general reward during the training process.

\subsection{Loss Function and Parameter Learning}
There are in total four sets of parameters to be learned. 
We denote all the parameters by $\Theta=\{{\mathbf{M}}, {\mathbf{w}}, w_p, \{\mathbf{p}_u\}\}$.
When dealing with real-world data, the training set consists of $(\mathcal{H}_i^u,\mathcal{A}_i^u,r_i^u)$, which records optimal actions provided by human, and hence the loss function is defined as follows:
\begin{eqnarray*}
\mathcal{L}(\Theta) = \mathbb{E}\left[\left(r_i^u+\gamma Q(\mathcal{H}_{i+1}^u, \mathcal{A}_{i+1}^u|\Theta) - Q(\mathcal{H}_i^u, \mathcal{A}_i^u|\Theta) \right)^2\right].
\end{eqnarray*}
In the on-policy training with a user simulator, the loss function is defined as
\begin{eqnarray*}
\mathcal{L}(\Theta) = \mathbb{E}\left[\left(r_i^u+\text{max}_{\mathcal{A}_{i+1}'}\gamma Q(\mathcal{H}_{i+1}^u, \mathcal{A}_{i+1}'|\Theta) - Q(\mathcal{H}_i^u, \mathcal{A}_i^u|\Theta) \right)^2\right],
\end{eqnarray*}
where $r_i^u$ is the reward obtained at time step $i$ and $\mathcal{H}_{i+1}^u$ is the update dialogue history at time step $i+1$. 

We use the value iteration method~\cite{bellman1957markovian} to learn both the general and personal Q-functions.
We adopt an online stochastic gradient descent algorithm \cite{bottou2010large} with learning rate $0.0001$ to optimize our model. Specifically, we use the State-Action-Reward-State-Action (SARSA) algorithm. In the on-policy training with the simulation, the model has decreasing probability $\eta=0.2 e^{-\frac{\beta}{1000}}$ of choosing a random reply in the candidate set so as to ensure the sufficient exploration, where $\beta$ is the number of training dialogues seen by the algorithm.

\subsection{Transfer Learning Algorithm}

\begin{wrapfigure}{r}{0.48\textwidth}
\vskip -0.3in
\begin{minipage}{.48\textwidth}
\scriptsize
\captionof{algorithm}{The PETAL Algorithm}\label{algorithm}
\begin{algorithmic}[1]
\State Input: $\mathcal{D}^s$,$\mathcal{D}^t$
\State Output: $\Theta=\{\mathbf{M}, \mathbf{w}, w_p\, \{ \mathbf{p}_u\} \}$
\Procedure{Transfer Algorithm}{$\mathcal{D}^s$,$\mathcal{D}^t$}
\State $\{\mathbf{M}, \mathbf{w}, w_p\} \gets$ \Call{Train-Source-Model}{$\mathcal{D}^s$}
\State $\{\mathbf{M}, \mathbf{w}, w_p, \{\mathbf{p}_u\}\}\gets$\Call{Transfer}{$\mathcal{D}^t$,$\mathbf{M}$,$\mathbf{w}$,$w_p$}
\EndProcedure

\Function{Train-Source-Model}{$\mathcal{D}^s$}
\For {$\{\mathcal{O}_i^{u}, \mathcal{A}_i^{u}\}$ in $\mathcal{D}^s$}
\For {${(\mathcal{H}_i^u,\mathcal{A}_i^u,r_i^u,\mathcal{H}_{i+1}^u,\mathcal{A}_{i+1}^u)}$ in $\{\mathcal{O}_i^{u}, \mathcal{A}_i^{u}\}$}
\State $\Theta_{t+1} \gets \Theta_t + \alpha \Delta_{\Theta} \mathcal{L}(\Theta_t) $
\EndFor
\EndFor
\Return $\{\mathbf{M}, \mathbf{w}, w_p\}$
\EndFunction

\Function{Transfer}{$\mathcal{D}^t$, $\mathbf{M}, \mathbf{w}, w_p$ }
\For {$\{\{\mathcal{O}_i^{u}, \mathcal{A}_i^{u}\}_i^T\}$ in $\mathcal{D}^t$}
\For {${(\mathcal{H}_i^u,\mathcal{A}_i^u,r_i^u,\mathcal{H}_{i+1}^u,\mathcal{A}_{i+1}^u)}$ in $\{\mathcal{O}_i^{u}, \mathcal{A}_i^{u}\}$}
\State $\Theta_{t+1} \gets \Theta_t + \alpha \Delta_{\Theta} \mathcal{L}(\Theta_t) $
\EndFor
\EndFor
\Return $\{\mathbf{M}, \mathbf{w}, w_p , \{ \mathbf{p}_u\} \}$
\EndFunction

\end{algorithmic}
\vskip -0.1in
\end{minipage}%
\vskip -0.1in
\end{wrapfigure}

The detailed PETAL algorithm is shown in Algorithm \ref{algorithm}.
We train our model for each user in the source domain. ${\mathbf{M}}$, ${\mathbf{w}}$ and $w_p$ are shared by all users and there is a separate ${\mathbf{p}_u}$ for each user in the source domain.
We transfer ${\mathbf{M}}$, ${\mathbf{w}}$ and $w_p$ to the target domain by using them to initialize the corresponding variables in the target domain, and then we train them as well as ${\mathbf{p}_u}$ for each target user with limited training data.
Since the source and target users might have different preferences, $\mathbf{p}_u$ learned in source domain is not very useful in the target domain. The personal preference of each target user will be learned separately in each ${\mathbf{p}_u}$. Without modelling ${\mathbf{p}_u}$ for each user, different preferences of source and target users might interfere with each other and thus cause negative transfer.

The number of parameters in our model is around $d^2+dv$, where $v$ is the total vocabulary size and $d$ is the dimension of the state vector. In our experiment where $v=1,500$ and $d=50$, the number of parameters in the general Q-function is about $85k$ and that for the personal Q-function is under $100$ for each user, hence the parameters in the personalized Q-function could be learned accurately with the limited data in the target domain.

\section{Experiments}
In this section, we experimentally verify the effectiveness of the proposed PETAL model by conducting experiments on a real-world dataset and a simulation dataset.
\subsection{Baselines}
We compare the proposed PETAL model with six baseline algorithms including ``NoneTL'' which is trained only with the data from target users, ``Sim''~\cite{casanueva2015knowledge} which is trained with the data from both target user and the most similar user in the source domain, ``Bandit''~\cite{genevay2016transfer} in which for each target user, the most useful source user is identified by a bandit algorithm, ``PriorSim''~\cite{gavsic2013pomdp} in which for each target user, the policy from the most similar user in the source domain is used as a prior, ``PriorAll''~\cite{gavsic2013pomdp} in which for each target user, the dialogue policy trained on all the users in the source domain is used as a prior, and ``All'' where the policy is trained on all source users' data.


\subsection{Experiments on Real-World Data}

\begin{wrapfigure}{r}{0.55\textwidth}
\vskip -0.2in
\begin{minipage}{0.55\textwidth}
\begin{table}[H]\small
\vspace{-4em}
\caption{Statistics of the datasets}
\begin{center}
\begin{tabular}{|c||c|c|c|c|}
  \hline
   & \multicolumn{2}{|c|}{Source Domain} & \multicolumn{2}{|c|}{Target Domain} \\
  \hline
  Dataset & Users & Dialogues & Users & Dialogues \\
  \hline
  Real Data & 52 & 1,859 & 20 & 329 \\
  \hline
  Simulation & 11 & 176,000 & 5 & 100 \\
  \hline
\end{tabular}
\label{tab:DataTab}
\end{center}
\end{table}
\end{minipage}
\vskip -0.2in
\end{wrapfigure}

In this section, we evaluate our model on a real-world dataset. This real-world dataset, which is collected between July 2015 and April 2016 from a O2O coffee ordering service, contains 2,185 coffee dialogues between 72 consumers and coffee makers. The users order coffee by providing the coffee type, the temperature, the cup size and the delivery address, hence there are 4 order choices. We select 52 users with more than 23 dialogues as the source domain. Each of the remaining 20 users is used separately as a target domain. In total, there are 1,859 coffee dialogues in the source domain and 329 coffee dialogues in the target domain. 221 earlier dialogues in the target domain are used as the training set and the remaining 108 dialogues form the test set. The statistics of this dataset is shown in Table~\ref{tab:DataTab}.


For each round of the testing conversation, a model will rank the ground truth reply $\mathcal{A}_i^u$ among 10 randomly chosen agent replies. The label assigned to $\mathcal{A}_i^u$ is 1 and those for randomly chosen agent replies are 0. By following \cite{williams2016end}, we calculate the AUC score for each turn in a conversation and the performance of an algorithm is measured by the average AUC score of each dialogue for every user in the test set.

\begin{figure}[!ht]
\centering\mbox{
\subfigure[{\small Real-world Average AUC (the higher the better)}]{\label{fig:ExpReal} \scalebox{0.45}{\includegraphics[width=\columnwidth]{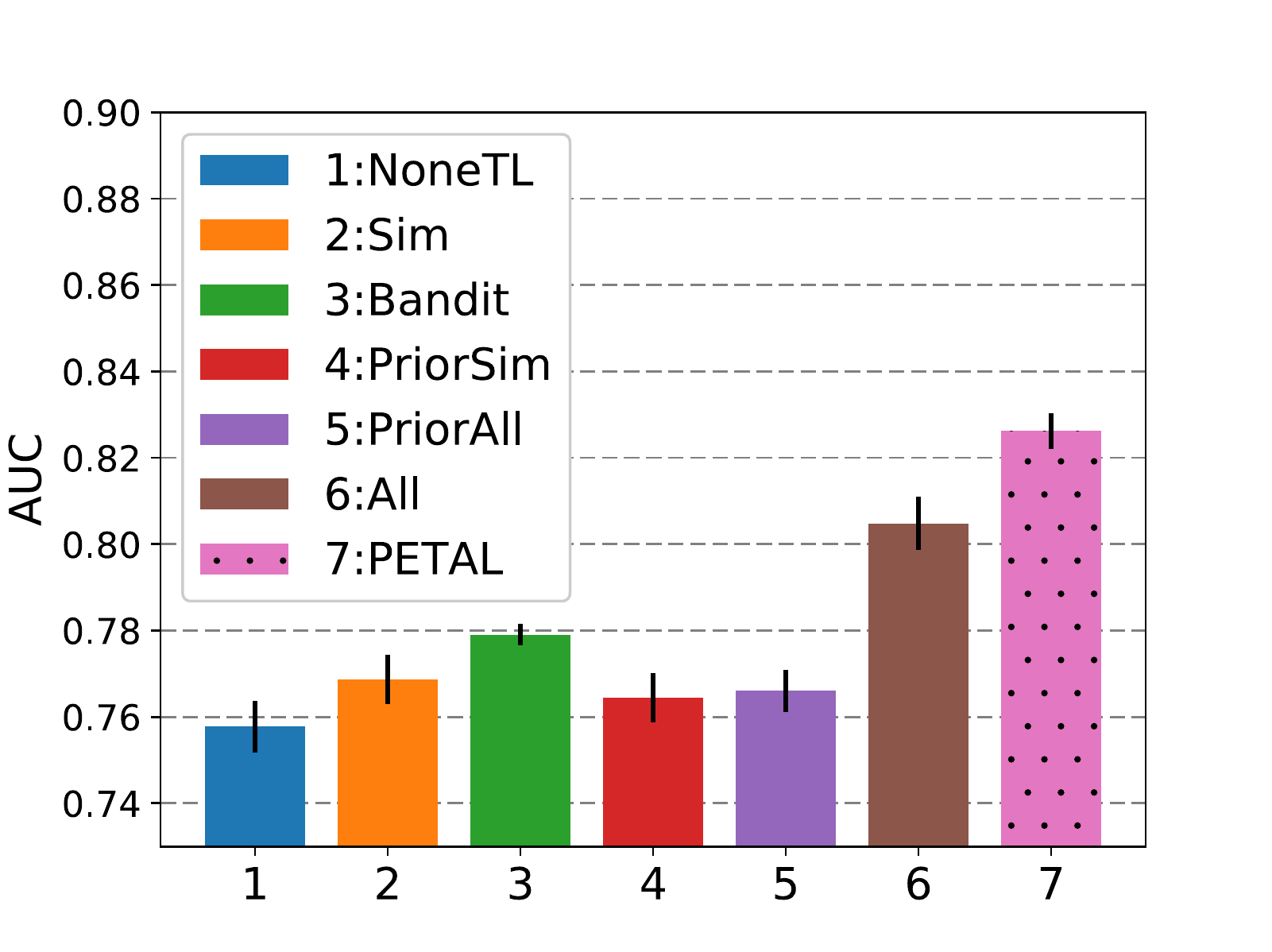}}}
\hfill
\subfigure[{\scriptsize Simulation Average Reward (the higher the better)}]{\label{fig:ExpSimReward} \scalebox{0.45}{\includegraphics[width=\columnwidth]{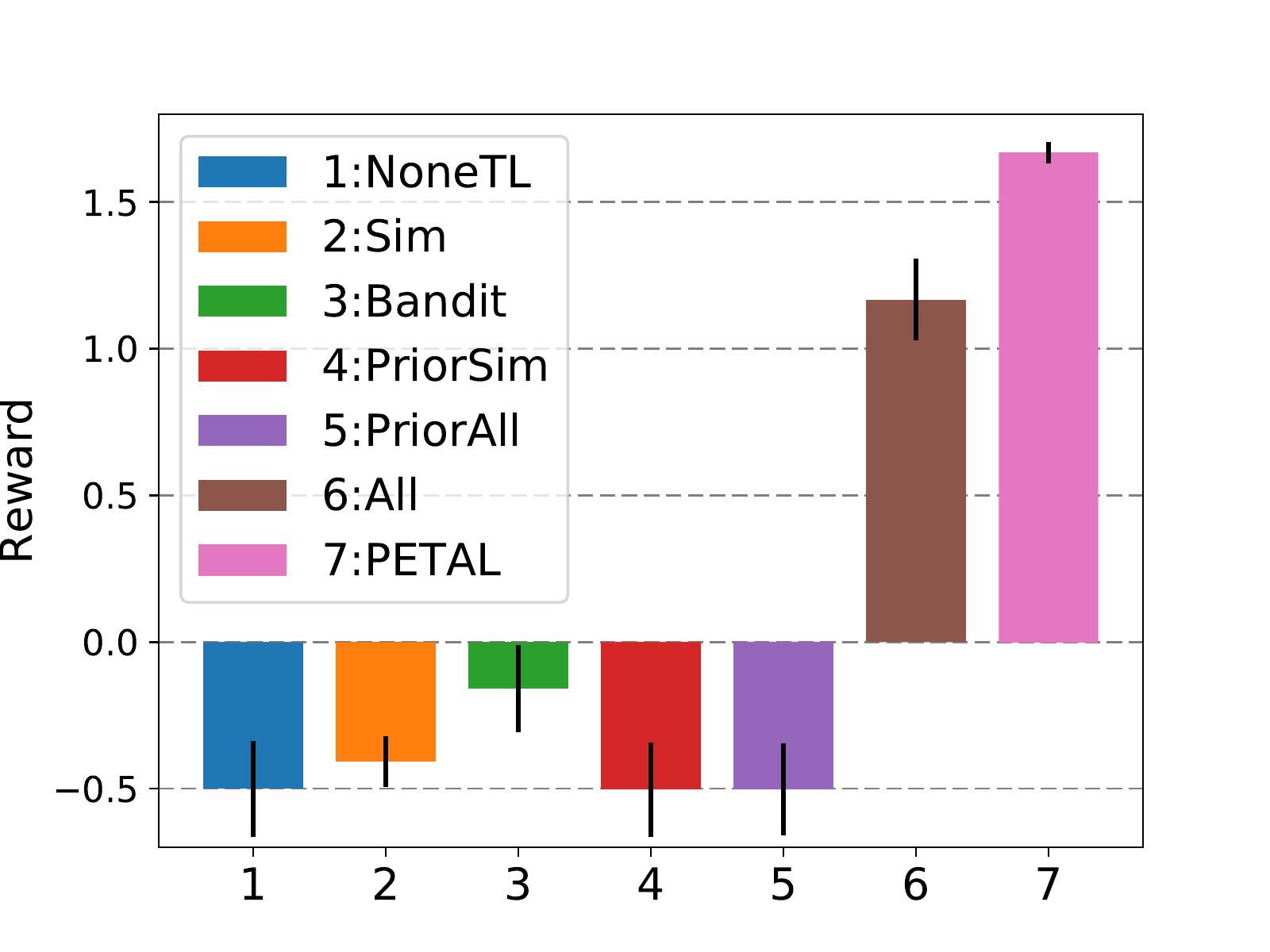}}}
\vspace{-1em}
}
\centering\mbox{\vspace{-2em}
\subfigure[{\small Simulation Success Rate (the higher the better)}]{\label{fig:ExpSimSuccess} \scalebox{0.45}{\includegraphics[width=\columnwidth]{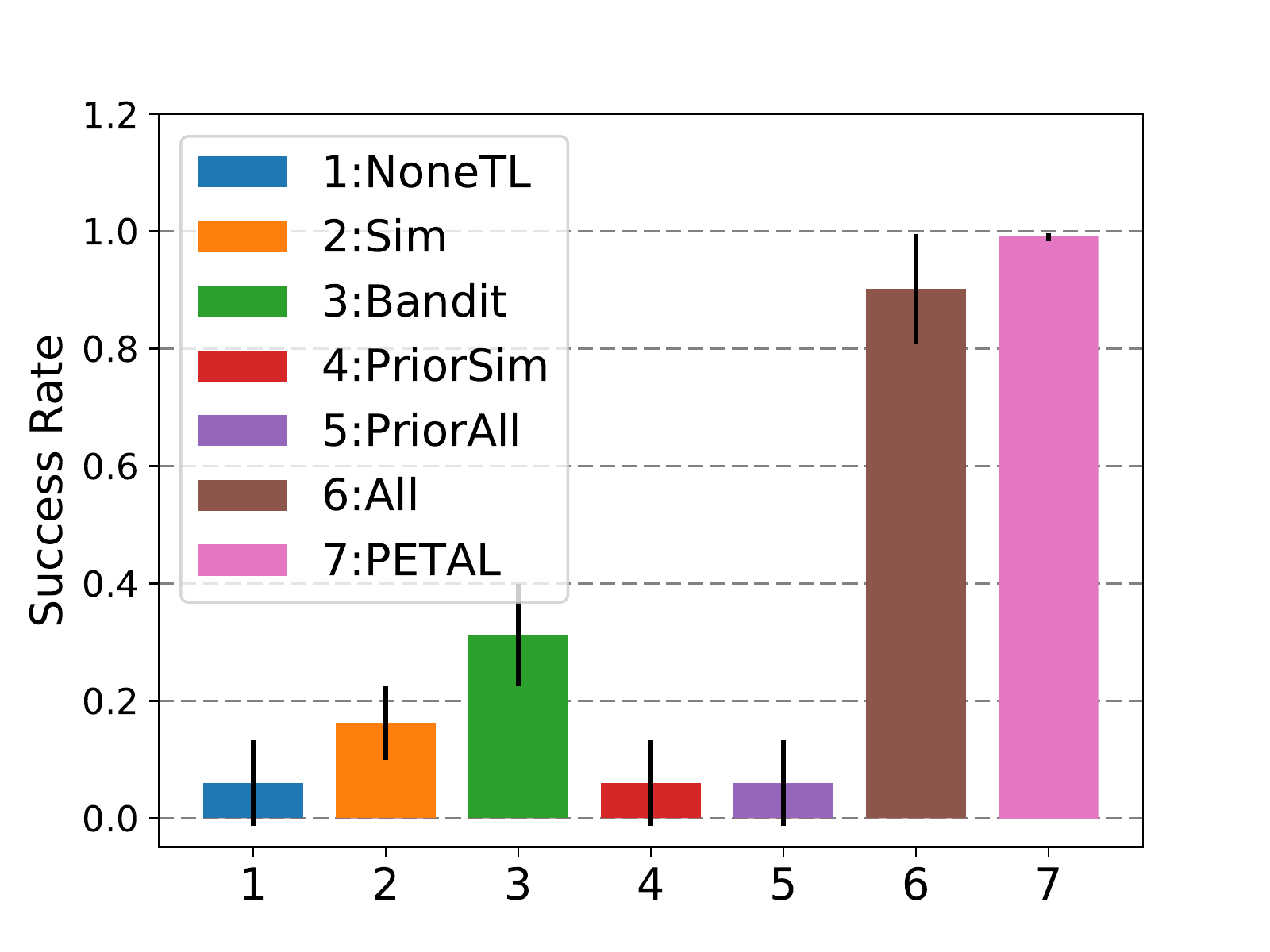}}}
\hfill
\subfigure[{\scriptsize Simulation Average Dialogue Length (the lower the better)}]{\label{fig:ExpSimLen} \scalebox{0.45}{\includegraphics[width=\columnwidth]{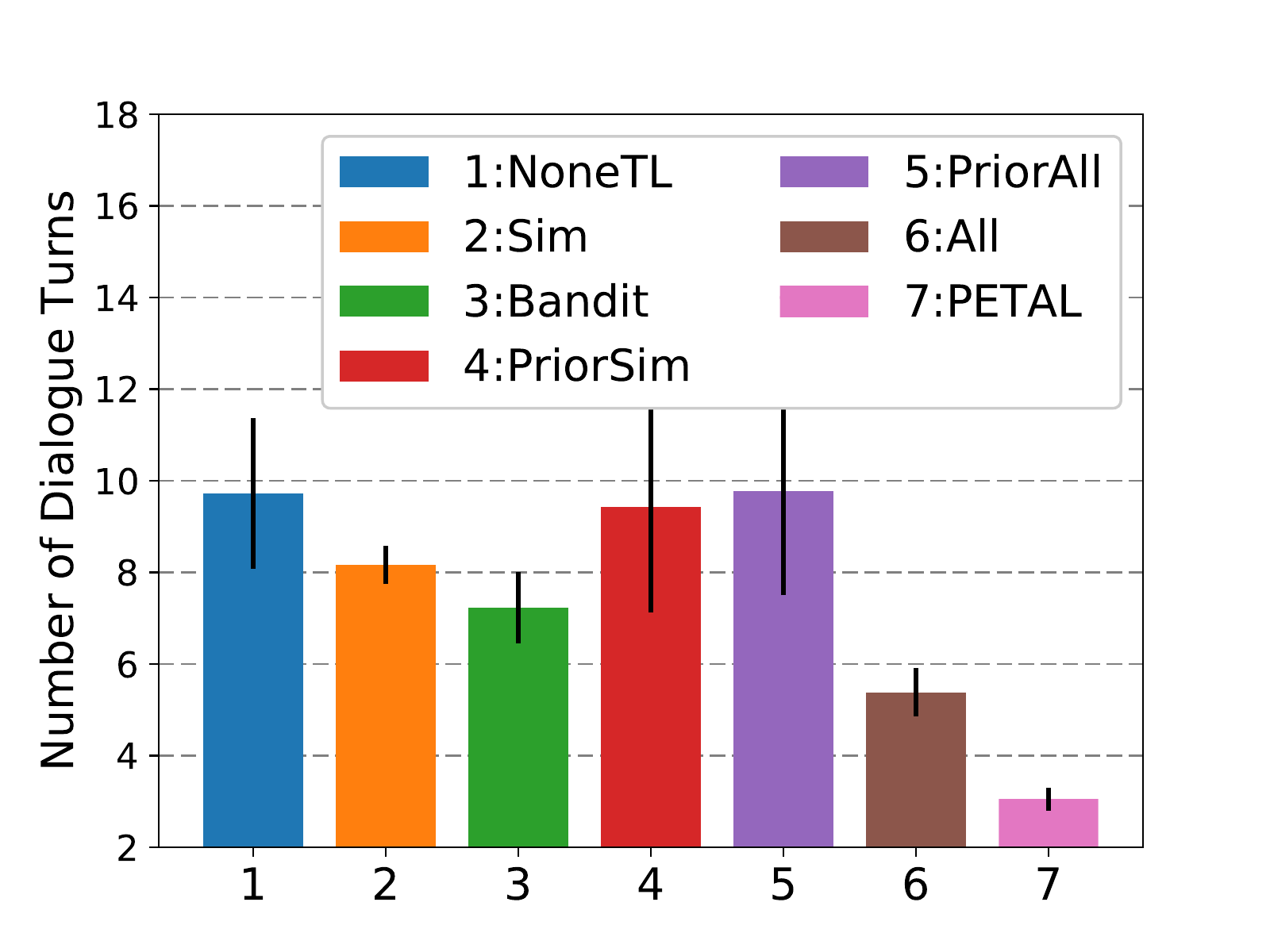}}}
}
\vspace{-1em}
\caption{Experimental results on real-world and simulations datasets.}
\vspace{-1em}
\end{figure}

In Figure~\ref{fig:ExpReal}, we report the mean and standard deviation of averaged AUC score with 5 different random \mbox{seeds}, which are used to randomly sample agent replies as candidates. The performance of ``NoneTL'', ``PriorSim'' and ``PriorAll'' are worse than ``All'' which directly transfers training data, because fitting only target domain data can cause the overfitting. Transferring data from similar users (i.e., ``Sim'') is not as good as transferring data from all source users (i.e., ``All''), because common knowledge has to be learned from more data. The proposed ``PETAL'' method performs the best because it learns common knowledge from all users and avoids the negative transfer caused by different preferences among source and target users, which indicates that the proposed personalized model fits dialogues better and demonstrates the effectiveness of PETAL on this real-world dataset.

\begin{wrapfigure}{r}{0.6\textwidth}
\begin{minipage}{0.6\textwidth}
\begin{table}[H]\small
\vspace{-4em}
\caption{A case study on the real-world dataset. The last column shows candidate responses, where the ground truth response is marked with *. The first and second columns show predicted rewards of ``All'' and ``PETAL'' on these candidates.}
\begin{center}
\begin{small}
	\begin{tabular}{rrl}
		  \hline
		  \multicolumn{2}{r}{User utterance}: & I want a cup of coffee. \\
		  \hline
		   All & PETAL & Response Candidates\\
		   0.86 & 1.36 & * Same as before? Tall hot americano and \\
		   & & deliver to Central Conservatory of Music?\\
		   0.99 & 0.92 & All right, deliver to No.1199 Beiyuan Road,\\
		   & & Chaoyang District, Beijing? \\
		   0.72 & 0.69 & What's your address? \\
		  \hline
	\end{tabular}
\label{tab:DialogueCaseReal}
\end{small}
\end{center}
\end{table}
\end{minipage}
\vskip -0.1in
\end{wrapfigure}

A case study is shown in Table~\ref{tab:DialogueCaseReal} and due to space limit, we only show three candidates.
From the results, we can see that the proposed ``PETAL'' method ranks the ground truth response in the first place based on the predicted reward given by the learned personalized Q-function but the ``All'' method without personalization ranks an wrong address higher, which demonstrates the effectiveness of the proposed method.

\subsection{Experiments on Simulation Data}
In this section, we compare our model with baseline models on the simulated coffee-ordering dialogue data. The simulated users order coffee by providing their coffee type, temperature, size and delivery address, and the agents reply by choosing from a set of predefined candidate responses without knowing the speech act.
We have 11 simulated users in the source domain, in which 10 users have their own coffee preferences while the rest one has no preference. The target domain has 5 users, which have different preferences from users in the source domain. A simulator is designed based on the real-world dataset used in the previous section. The simulator will order according to his preference with probability $0.8$ and otherwise the simulator will order coffee randomly. The training set of each user in the target domain has 20 dialogues and the test set has 300 dialogues.
The reward in the experiment is the same as the reward defined in Section \ref{sec:reward}. 

Each model will choose a reply from a set of candidates generated with templates at each turn, and the simulator will react to the reply accordingly. For each model, we report the mean and standard deviation of averaged reward \cite{genevay2016transfer}, averaged success rate \cite{casanueva2015knowledge} and averaged dialogue length over all possible target users, repeated for 5 times with different random seeds.

The results are shown in
Figure~\ref{fig:ExpSimReward},
Figure~\ref{fig:ExpSimSuccess} and
Figure~\ref{fig:ExpSimLen}. PETAL outperforms all baselines and obtains the highest average reward, the highest success rate and the lowest dialogue length, which implies that PETAL has found a better dialogue policy which can adapt its behaviour according to the preference of target users and again demonstrates the effectiveness of PETAL in a live environment.

We show a typical case for the simulation data in Tables~\ref{tab:DialogCaseTable} and \ref{tab:DialogCaseTable2}. The non-personalized dialogue system corresponding to the ``All'' model has to ask the users all the choices even for frequent users in Table \ref{tab:DialogCaseTable2}, because there is no universal recommendation for all the frequent users with different preferences. However, PETAL has learned the target users' preferences in previous dialogues. As shown in Table~\ref{tab:DialogCaseTable}, the response from the agent is specially tailored for the target user because personalized questions given by the PETAL method can guide the user to complete the coffee-ordering task faster than general questions, leading to shorter dialogue and higher averaged reward. If the user does not want everything as usual, which is shown in the second case of Table \ref{tab:DialogCaseTable}, PETAL can still react correctly due to the shared dialogue knowledge transferred from the source domain. These cases show that PETAL can choose different optimal actions for different users and effectively shorten the conversation.

\begin{table}
\begin{center}
\begin{scriptsize}
    \begin{minipage}[t]{0.48\hsize}\centering
    \caption{Personalized Dialogue Cases}
		\begin{tabular}{rl}
		  \hline
		  User: & I want a cup of coffee. \\
		  Agent: & Same as before? Tall hot macchiato and deliver to\\
		  & No.1199 Minsheng Road, Pudong District Shanghai?\\
		  User: & Yes. \\
		  Agent: & Please pay. \\
		  User: & Payment completed. \\
		  \hline
		  User: & I want a cup of coffee. \\
		  Agent: & Same as before? Tall hot macchiato and deliver to\\
		  & No.1199  Minsheng Road, Pudong District, Shanghai? \\
		  User: & I want iced mocha today. \\
		  Agent: & Sure, please pay. \\
		  User: & Payment completed. \\
		  \hline
		\end{tabular}
        \label{tab:DialogCaseTable}
    \end{minipage}
    \hfill
    \begin{minipage}[t]{0.48\hsize}\centering
    \caption{A Non-Personalized Dialogue Case}
		\begin{tabular}{rl}
		  \hline
		  User: & I want a cup of coffee. \\
		  Agent: & Hot coffee? \\
		  User: & Yes. \\
		  Agent: & What would you like to drink? \\
		  User: & Hot macchiato, please. \\
		  Agent: & What's your address? \\
		  User: & No.1199 Minsheng Road, Pudong District, Shanghai.\\
		  Agent: & Alright. Tall, Grande, or Venti? \\
		  User: & Tall. \\
		  Agent: & Please pay. \\
		  User: & Payment completed. \\
		  \hline
		\end{tabular}
        \label{tab:DialogCaseTable2}
    \end{minipage}
\end{scriptsize}
\end{center}
\vspace{-2em}
\end{table}

\section{Conclusion}
In this paper, we tackle the problem of designing a personalized dialogue system. We propose the PETAL system, a transfer learning framework based on the POMDP, for learning a personalized dialogue system. The PETAL system first learns common dialogue knowledge from the source domain and then adapts this knowledge to the target user.
We propose to model a personalized policy in the POMDP with a personalized Q-function. This framework can avoid the negative transfer problem brought by differences between the source users and the target user.
Experimental results on the real-world coffee-ordering data and the simulation data show that PETAL can learn different optimal policies for different users, and thus effectively improve the dialogue quality under the personalized setting.
As a future direction, we will investigate to transfer knowledge from heterogeneous domains such as knowledge graphs and images.

\bibliographystyle{plain}
\bibliography{PETAL}

\begin{thebibliography}{27}
\providecommand{\natexlab}[1]{#1}
\providecommand{\url}[1]{\texttt{#1}}
\expandafter\ifx\csname urlstyle\endcsname\relax
  \providecommand{\doi}[1]{doi: #1}\else
  \providecommand{\doi}{doi: \begingroup \urlstyle{rm}\Url}\fi

\bibitem[Bang et~al.(2015)Bang, Noh, Kim, and Lee]{bang2015example}
Jeesoo Bang, Hyungjong Noh, Yonghee Kim, and Gary~Geunbae Lee.
\newblock Example-based chat-oriented dialogue system with personalized
  long-term memory.
\newblock In \emph{Proceedings of International Conference on Big Data and
  Smart Computing}, pages 238--243, 2015.

\bibitem[Bellman(1957)]{bellman1957markovian}
Richard Bellman.
\newblock A {M}arkovian decision process.
\newblock Technical report, DTIC Document, 1957.

\bibitem[Bottou(2010)]{bottou2010large}
L{\'e}on Bottou.
\newblock Large-scale machine learning with stochastic gradient descent.
\newblock In \emph{Proceedings of 19th International Conference on
  Computational Statistics}, pages 177--186, 2010.

\bibitem[Casanueva et~al.(2015)Casanueva, Hain, Christensen, Marxer, and
  Green]{casanueva2015knowledge}
Inigo Casanueva, Thomas Hain, Heidi Christensen, Ricard Marxer, and Phil Green.
\newblock Knowledge transfer between speakers for personalised dialogue
  management.
\newblock In \emph{Proceedings of the 16th Annual Meeting of the Special
  Interest Group on Discourse and Dialogue}, 2015.

\bibitem[Galley et~al.(2015)Galley, Brockett, Sordoni, Ji, Auli, Quirk,
  Mitchell, Gao, and Dolan]{galley2015deltableu}
Michel Galley, Chris Brockett, Alessandro Sordoni, Yangfeng Ji, Michael Auli,
  Chris Quirk, Margaret Mitchell, Jianfeng Gao, and Bill Dolan.
\newblock {deltaBLEU}: A discriminative metric for generation tasks with
  intrinsically diverse targets.
\newblock \emph{arXiv preprint arXiv:1506.06863}, 2015.

\bibitem[Ga{\v{s}}ic et~al.(2013)Ga{\v{s}}ic, Breslin, Henderson, Kim, Szummer,
  Thomson, Tsiakoulis, and Young]{gavsic2013pomdp}
Milica Ga{\v{s}}ic, Catherine Breslin, Matthew Henderson, Dongho Kim, Martin
  Szummer, Blaise Thomson, Pirros Tsiakoulis, and Steve Young.
\newblock {POMDP}-based dialogue manager adaptation to extended domains.
\newblock In \emph{Proceedings of the 14th Annual Meeting of the Special
  Interest Group on Discourse and Dialogue}, 2013.

\bibitem[Gasic et~al.(2014)Gasic, Kim, Tsiakoulis, Breslin, Henderson, Szummer,
  Thomson, and Young]{gavsic2014incremental}
Milica Gasic, Dongho Kim, Pirros Tsiakoulis, Catherine Breslin, Matthew
  Henderson, Martin Szummer, Blaise Thomson, and Steve~J. Young.
\newblock Incremental on-line adaptation of {POMDP}-based dialogue managers to
  extended domains.
\newblock In \emph{Proceedings of the 15th Annual Conference of the
  International Speech Communication Association}, pages 140--144, 2014.

\bibitem[Genevay and Laroche(2016)]{genevay2016transfer}
Aude Genevay and Romain Laroche.
\newblock Transfer learning for user adaptation in spoken dialogue systems.
\newblock In \emph{Proceedings of the 2016 International Conference on
  Autonomous Agents \& Multiagent Systems}, pages 975--983, 2016.

\bibitem[Hiraoka et~al.(2014)Hiraoka, Neubig, Sakti, Toda, and
  Nakamura]{hiraoka2014reinforcement}
Takuya Hiraoka, Graham Neubig, Sakriani Sakti, Tomoki Toda, and Satoshi
  Nakamura.
\newblock Reinforcement learning of cooperative persuasive dialogue policies
  using framing.
\newblock In \emph{Proceedings of the 25th International Conference on
  Computational Linguistics}, pages 1706--1717, 2014.

\bibitem[Kim et~al.(2014)Kim, Bang, Choi, Ryu, Koo, and
  Lee]{kim2014acquisition}
Yonghee Kim, Jeesoo Bang, Junhwi Choi, Seonghan Ryu, Sangjun Koo, and
  Gary~Geunbae Lee.
\newblock Acquisition and use of long-term memory for personalized dialog
  systems.
\newblock In \emph{Proceedings of International Workshop on Multimodal Analyses
  Enabling Artificial Agents in Human-Machine Interaction}, pages 78--87, 2014.

\bibitem[Levin et~al.(1997)Levin, Pieraccini, and Eckert]{levin1997learning}
Esther Levin, Roberto Pieraccini, and Wieland Eckert.
\newblock Learning dialogue strategies within the {M}arkov decision process
  framework.
\newblock In \emph{Proceedings of IEEE Workshop on Automatic Speech Recognition
  and Understanding}, pages 72--79, 1997.

\bibitem[Li et~al.(2016)Li, Monroe, Ritter, and Jurafsky]{li2016deep}
Jiwei Li, Will Monroe, Alan Ritter, and Dan Jurafsky.
\newblock Deep reinforcement learning for dialogue generation.
\newblock \emph{arXiv preprint arXiv:1606.01541}, 2016.

\bibitem[Mou et~al.(2016)Mou, Song, Yan, Li, Zhang, and Jin]{mou2016sequence}
Lili Mou, Yiping Song, Rui Yan, Ge~Li, Lu~Zhang, and Zhi Jin.
\newblock Sequence to backward and forward sequences: A content-introducing
  approach to generative short-text conversation.
\newblock \emph{arXiv preprint arXiv:1607.00970}, 2016.

\bibitem[Pan and Yang(2010)]{pan2010survey}
Sinno~Jialin Pan and Qiang Yang.
\newblock A survey on transfer learning.
\newblock \emph{IEEE Transactions on Knowledge and Data Engineering},
  22\penalty0 (10):\penalty0 1345--1359, 2010.

\bibitem[Ritter et~al.(2011)Ritter, Cherry, and Dolan]{ritter2011data}
Alan Ritter, Colin Cherry, and William~B Dolan.
\newblock Data-driven response generation in social media.
\newblock In \emph{Proceedings of the 2011 Conference on Empirical Methods in
  Natural Language Processing}, pages 583--593, 2011.

\bibitem[Rosenfeld and Kraus(2016{\natexlab{a}})]{rosenfeld2016providing}
Ariel Rosenfeld and Sarit Kraus.
\newblock Providing arguments in discussions on the basis of the prediction of
  human argumentative behavior.
\newblock \emph{ACM Transactions on Interactive Intelligent Systems},
  6\penalty0 (4):\penalty0 30, 2016{\natexlab{a}}.

\bibitem[Rosenfeld and Kraus(2016{\natexlab{b}})]{rosenfeld2016strategical}
Ariel Rosenfeld and Sarit Kraus.
\newblock Strategical argumentative agent for human persuasion.
\newblock In \emph{Proceedings of the 22nd European Conference on Artificial
  Intelligence}, 2016{\natexlab{b}}.

\bibitem[Serban et~al.(2015)Serban, Sordoni, Bengio, Courville, and
  Pineau]{serban2015hierarchical}
Iulian~V Serban, Alessandro Sordoni, Yoshua Bengio, Aaron Courville, and Joelle
  Pineau.
\newblock Hierarchical neural network generative models for movie dialogues.
\newblock \emph{arXiv preprint arXiv:1507.04808}, 2015.

\bibitem[Tan et~al.(2014)Tan, Zhong, Xiang, and Yang]{tan2014multi}
Ben Tan, Erheng Zhong, Evan~Wei Xiang, and Qiang Yang.
\newblock Multi-transfer: Transfer learning with multiple views and multiple
  sources.
\newblock \emph{Statistical Analysis and Data Mining}, 7\penalty0 (4):\penalty0
  282--293, 2014.

\bibitem[Tan et~al.(2015)Tan, Song, Zhong, and Yang]{tan2015transitive}
Ben Tan, Yangqiu Song, Erheng Zhong, and Qiang Yang.
\newblock Transitive transfer learning.
\newblock In \emph{Proceedings of the 21th ACM SIGKDD International Conference
  on Knowledge Discovery and Data Mining}, pages 1155--1164, 2015.

\bibitem[Taylor and Stone(2009)]{taylor2009transfer}
Matthew~E Taylor and Peter Stone.
\newblock Transfer learning for reinforcement learning domains: A survey.
\newblock \emph{Journal of Machine Learning Research}, 10:\penalty0 1633--1685,
  2009.

\bibitem[Thompson et~al.(2004)Thompson, Goker, and
  Langley]{thompson2004personalized}
Cynthia~A Thompson, Mehmet~H Goker, and Pat Langley.
\newblock A personalized system for conversational recommendations.
\newblock \emph{Journal of Artificial Intelligence Research}, 21:\penalty0
  393--428, 2004.

\bibitem[Wei et~al.(2016)Wei, Zheng, and Yang]{wei2016transfer}
Ying Wei, Yu~Zheng, and Qiang Yang.
\newblock Transfer knowledge between cities.
\newblock In \emph{Proceedings of the 22nd ACM SIGKDD International Conference
  on Knowledge Discovery and Data Mining}, pages 1905--1914, 2016.

\bibitem[Wen et~al.(2015)Wen, Gasic, Mrksic, Su, Vandyke, and
  Young]{wen2015semantically}
Tsung-Hsien Wen, Milica Gasic, Nikola Mrksic, Pei-Hao Su, David Vandyke, and
  Steve Young.
\newblock Semantically conditioned lstm-based natural language generation for
  spoken dialogue systems.
\newblock \emph{arXiv preprint arXiv:1508.01745}, 2015.

\bibitem[Wen et~al.(2016)Wen, Gasic, Mrksic, Rojas-Barahona, Su, Ultes,
  Vandyke, and Young]{wen2016network}
Tsung-Hsien Wen, Milica Gasic, Nikola Mrksic, Lina~M Rojas-Barahona, Pei-Hao
  Su, Stefan Ultes, David Vandyke, and Steve Young.
\newblock A network-based end-to-end trainable task-oriented dialogue system.
\newblock \emph{arXiv preprint arXiv:1604.04562}, 2016.

\bibitem[Williams and Zweig(2016)]{williams2016end}
Jason~D Williams and Geoffrey Zweig.
\newblock End-to-end {LSTM}-based dialog control optimized with supervised and
  reinforcement learning.
\newblock \emph{arXiv preprint arXiv:1606.01269}, 2016.

\bibitem[Young et~al.(2013)Young, Ga{\v{s}}i{\'c}, Thomson, and
  Williams]{young2013pomdp}
Steve Young, Milica Ga{\v{s}}i{\'c}, Blaise Thomson, and Jason~D Williams.
\newblock {POMDP}-based statistical spoken dialog systems: A review.
\newblock \emph{Proceedings of the IEEE}, 101\penalty0 (5):\penalty0
  1160--1179, 2013.

\end{thebibliography}

\end{document}